\documentclass[twoside,leqno,twocolumn]{article}

\usepackage[letterpaper]{geometry}

\usepackage{ltexpprt}
\usepackage{hyperref}
\usepackage{pifont}
\usepackage{multirow}
\usepackage{amsmath,amssymb,amsfonts}
\usepackage{booktabs}
\usepackage[most]{tcolorbox}
\newtcolorbox{promptbox}{
colback=gray!5,  
colframe=black!75, 
left=1em, 
right=1em, 
top=1em, 
bottom=1em, 
sharp corners, 
boxrule=1pt 
}
\usepackage{algorithm}
\usepackage{algorithmic}
\usepackage{caption}
\begin{document}

\newcommand\relatedversion{}
\renewcommand\relatedversion{\thanks{The full version of the paper can be accessed at \protect\url{https://arxiv.org/abs/1902.09310}}} 

\title{\Large Disinformation Detection: An Evolving Challenge in the Age of LLMs
}
\author{Bohan Jiang\thanks{School of Computing and AI, Arizona State University, Tempe, AZ. \{bjiang14, ztan36, anirmal1, huanliu\}@asu.edu}
\and Zhen Tan\footnotemark[1]
\and Ayushi Nirmal\footnotemark[1]
\and Huan Liu\footnotemark[1]}


\date{}

\maketitle


\fancyfoot[R]{\scriptsize{Copyright \textcopyright\ 2024 by SIAM\\
Unauthorized reproduction of this article is prohibited}}





\begin{abstract} \small\baselineskip=9pt {The advent of generative Large Language Models (LLMs) such as ChatGPT has catalyzed transformative advancements across multiple domains. However, alongside these advancements, they have also introduced potential threats. One critical concern is the misuse of LLMs by disinformation spreaders, leveraging these models to generate highly persuasive yet misleading content that challenges the disinformation detection system. This work aims to address this issue by answering three research questions: (1) To what extent can the current disinformation detection technique reliably detect LLM-generated disinformation? (2) If traditional techniques prove less effective, can LLMs themself be exploited to serve as a robust defense against advanced disinformation? and, (3) Should both these strategies falter, what novel approaches can be proposed to counter this burgeoning threat effectively? 
A holistic exploration for the formation and detection of disinformation is conducted to foster this line of research.}
\end{abstract}




\section{Introduction}
\label{sec:Introduction}
The rise of Large Language Lodels (LLMs), exemplified by models such as ChatGPT~\cite{OpenAI2023GPT4TR} and Llama~\cite{touvron2023llama}, been a significant milestone in the domain of Computational Social Science (CSS). While LLMs have paved the way for expansive studies of human language and behavior~\cite{ziems2023can}, a pressing concern is their potential for misuse such as disinformation generation and propagation. As these models evolve in their capacity to generate increasingly persuasive human-level content, there exists a concomitant risk of their deployment in intentionally generating misleading information at scale. A concerning remark from a recent study underscores this~\cite{spitale2023ai} --- \textit{``the fact that AI-generated disinformation is not only cheaper and faster, but also more effective, gives me nightmares.''}

In the era preceding LLMs, research in AI-generated disinformation detection predominantly revolved around relatively Smaller Language Models (SLMs) such as BERT~\cite{devlin2018bert}, GPT-2~\cite{brown2020language}, and T5~\cite{raffel2020exploring}. The advent of LLMs, with their billion-scale parameters, has dramatically escalated the complexity of disinformation detection. The textual content generated by these LLMs is natural and human-sounding~\cite{soni2023comparing}. This evolution raises critical questions about the robustness and adaptability of current disinformation detection techniques, which were primarily designed around SLMs. Despite the significance, the consequences of this shift have not been extensively studied. This paper is motivated to bridge this knowledge gap by answering the following research questions:

\begin{figure}[!t]
\centering
\includegraphics[width=1\columnwidth]{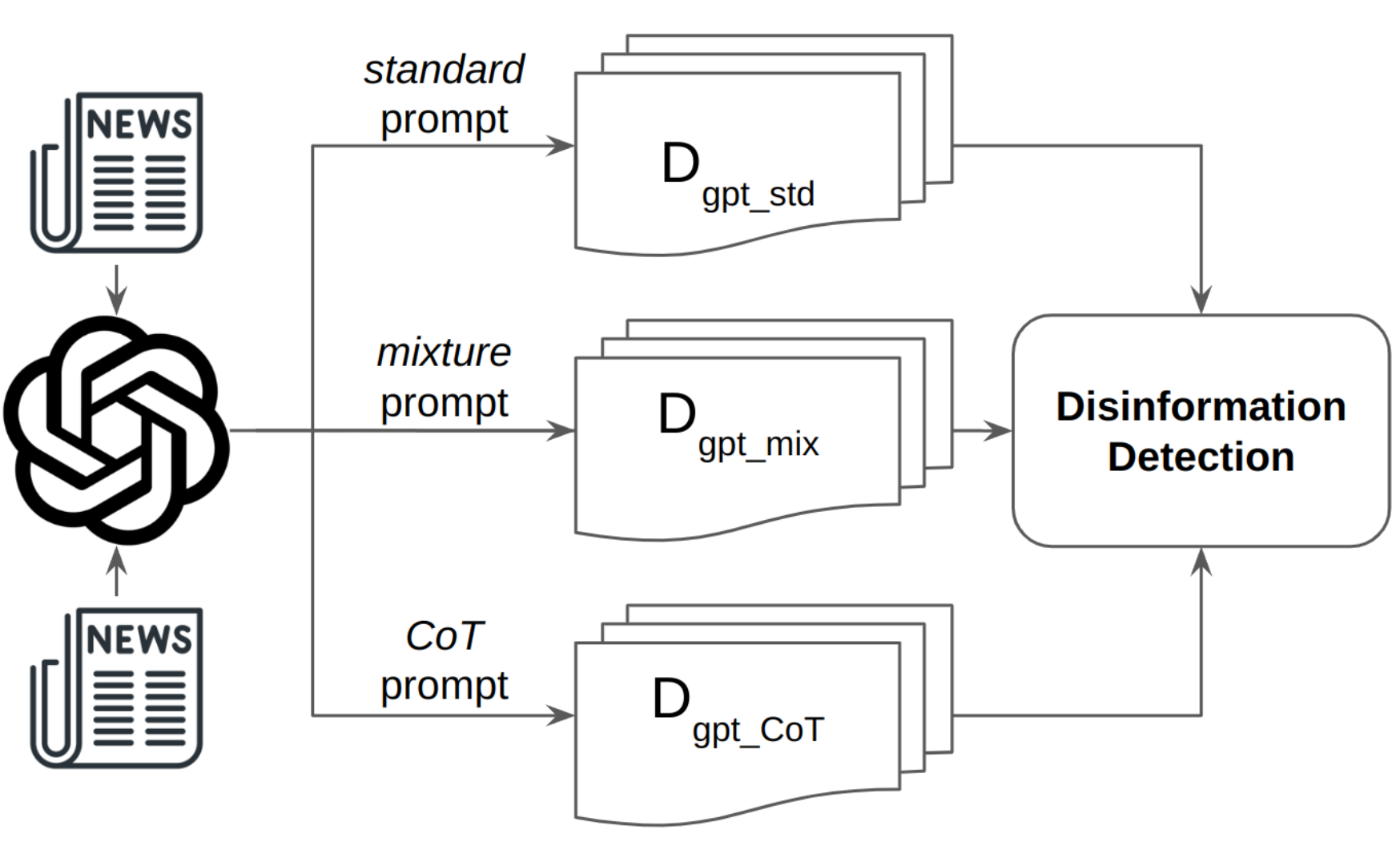} 
\caption{Overview of disinformation generation and detection using ChatGPT. We input human-crafted disinformation (left) along with distinct prompts to produce three separate LLM-generated disinformation datasets (center). We subsequently evaluate the efficacy of the disinformation detection system~(right) against LLM-generated disinformation. Detailed descriptions for the different types of prompts are elaborated later.}
\label{fig:overview}
\end{figure}

\begin{itemize}
    \item \textbf{RQ1:} Are existing disinformation detection techniques apt for LLM-generated disinformation?
    \item \textbf{RQ2:} If not, can LLMs themselves be adapted to detect such disinformation?
    \item \textbf{RQ3:} If both avenues fall short, what alternative solutions can be considered?
\end{itemize}

To ensure our findings are grounded in practical implications,  we contextualize our research within a real-world scenario: supposing a scenario wherein a malicious actor intends to leverage LLMs to generate ``advanced'' disinformation with the goal of fooling automated detection systems and swaying public perception. To this end, in this study, we start with a widely-used benchmark dataset~\cite{ahmed2017detection} comprising human-written news articles that are categorized as either \textit{fake} or \textit{true}. Based on this dataset, we construct novel disinformation datasets ($D_\text{gpt\_std}$, $D_\text{gpt\_mix}$, and $D_\text{gpt\_cot}$) of varying complexity levels with ChatGPT (GPT-3.5 and 4) using three prompt techniques. A high-level overview of our disinformation generation and detection is presented in Figure~\ref{fig:overview}. Addressing \textbf{RQ1}, we initially employ a state-of-the-art disinformation detection method~\cite{iceland2023good}, which involves fine-tuning a RoBERTa-based model~\cite{liu2019roberta} on human-written disinformation datasets ($D_\text{human}$). Subsequently, we evaluate the effectiveness of the fine-tuned RoBERTa-based model in detecting LLM-generated disinformation. For \textbf{RQ2}, we turn the lens towards LLMs themself, probing their ability to discern the self-generated disinformation. Lastly, for \textbf{RQ3}, we propose an innovative promoting method that aspires to emulate the human fact-checking process, leveraging LLMs to effectively detect advanced disinformation.

Through comprehensive experiments, we obtained several crucial observations. \ding{182}~Initial analyses indicate that while the fine-tuned RoBERTa model can accurately detect ``simple'' LLM-generated disinformation, it fails when confronted with disinformation of higher difficulty level generated from ``advanced'' prompts. Notably, for disinformation generated using chain-of-thought (CoT) prompts as detailed by~\cite{wei2022chain}, the fine-tuned detection model has an alarmingly high misclassification rate of 77.9\%. Particularly, a more detailed examination reveals a discernible political bias in the detection model. Our results demonstrate that while the model exhibits a pronounced inclination towards categorizing center-leaning news as \textit{true}, it tends to classify liberal- and conservative-leaning narratives as \textit{fake}. \ding{183}~Furthermore, we observe that vanilla ChatGPT cannot effectively detect disinformation generated even by itself. \ding{184}~However, our research also unveils an avenue for improvement: By leveraging a carefully designed CoT-inspired prompt, we can significantly increase the detection accuracy. In summary, this work has the following contributions: 

\begin{itemize}
    \item \underline{\textit{Dataset Curation}}: We construct three LLM-generated datasets to facilitate the area of disinformation detection;
    \item \underline{\textit{Problem Validation}}: In contrast to previous work, we demonstrate that existing detection techniques (SLMs) cannot effectively detect LLM-generated disinformation; 
    \item \underline{\textit{Framework Proposed}}: We propose novel methods for LLM-generated disinformation detection.
\end{itemize}

\section{Related Work}
Our focus in this paper is twofold: i) disinformation detection; and ii) text generation using LLMs. We now discuss the related literature across them.

\subsection{Disinformation Detection}
While \textit{misinformation} refers to false or inaccurate information that is spread without necessarily having the intent to deceive, \textit{disinformation} is deliberately fabricated or manipulated information intended to deceive or mislead people. They belong to the family of fake news~\cite{shu2017fake}. The inundation of misleading content in today's digital age surpasses the capacities of conventional manual fact-checking approaches, compelling the pursuit of automated countermeasures. In response, scholars have gravitated towards advanced computational methodologies for the automated detection of disinformation~\cite{nasir2021fake}.

In recent years, an important milestone in disinformation detection has been the development of deep learning. People train deep neural networks or on a large corpus to learn various textual features such as semantic meaning, writing styles, and tonal subtleties~\cite{zhang2015character}. For instance, Ruchansky et al.~\cite{ruchansky2017csi} introduced the CSI model, a hybrid deep learning model fusing content, social, and temporal information to enhance fake news detection. FakeBERT~\cite{kaliyar2021fakebert}, a BERT-based model that amalgamates the several blocks of a single-layer convolutional neural network (CNN), endowed with varied kernel dimensions and filters, alongside BERT. However, there remains a gap in the literature addressing the detection of disinformation generated by LLMs. A recent work provides an initial exploration into the effects of AI-generated COVID-19 misinformation on both detection systems and human perception~\cite{zhou2023synthetic}.~\cite{spitale2023ai} posited that disinformation generated by GPT-3 is often indistinguishable from that crafted by humans.

Our study aims to address this challenge, concentrating explicitly on LLM-generated disinformation and assessing the robustness of current detection methods when faced with this novel challenge.

\subsection{Large Language Models for Text Generation}
In recent years, the natural language processing (NLP) community has witnessed a shift in language models from SLMs with millions of parameters to the emergence of LLMs boasting billions of parameters. This transition has yielded significant advancements in various text generation tasks~\cite{zhou2023comprehensive}. Notably, models such as LaMDA~\cite{thoppilan2022lamda} with its impressive 137 billion parameters, OPT's 175 billion parameters~\cite{zhang2022opt}, Bloom's 176 billion parameters~\cite{scao2022bloom}, and PaLM's 540 billion parameters~\cite{chowdhery2022palm}, alongside the popular GPT family (including GPT-3, 3.5, and 4)~\cite{koubaa2023gpt}, have shown increasingly ability to generate human-level responses based on the input few-shot or zero-shot~\cite{qin2023chatgpt}. However, it is essential to acknowledge that the format of the input prompt can impact the performance~\cite{white2023prompt}. Leveraging advanced prompt engineering techniques, such as those explored in recent research~\cite{wei2022chain}, can effectively guide LLMs to produce responses that are not only more accurate but also of higher quality.

Trained on a large online corpus, ChatGPT is a repository of diverse knowledge. What sets ChatGPT apart is its unique training methodology --- Reinforcement Learning with Human Feedback (RLHF)~\cite{ouyang2022training}. In this way, human feedback is systematically incorporated into generating and selecting optimal results. Moreover, ChatGPT is accessible to the public through OpenAI APIs and a concise online chatbot interface. Given these features, our work harnesses ChatGPT (GPT-3.5 and 4) to generate ``high-quality'' disinformation.

\section{Datasets}
In this section, we introduce the data collection process. We begin with a human-written fake news dataset (i.e., $D_{human}$). Based on it, we build three LLM-generated fake news datasets (i.e., $D_\text{gpt\_std}$, $D_\text{gpt\_mix}$, and $D_\text{gpt\_cot}$) upon $D_\text{human}$ using distinct zero-shot prompt techniques. We contribute these datasets as novel resources for facilitating future research in the detection of LLM-generated disinformation. Statistics of our datasets are provided in Table~\ref{tab:compare}. 

\begin{table}[!t]
    \centering
    \caption{Datasets statistics}
    \scalebox{0.9}{
    \begin{tabular}{l|c|c|c|c}
         \toprule
         {}&\text{$D_\text{human}$} & \text{$D_\text{gpt\_std}$} & \text{$D_\text{gpt\_mix}$} & {$D_\text{gpt\_cot}$} \\
         \midrule
         \# of samples & 23,525 & 23,278 & 1,000 & 1,737\\ \hline 
         headline & \checkmark & \checkmark &  &  \\ \hline
         content & \checkmark & \checkmark & \checkmark & \checkmark \\

         \bottomrule \hline
    \end{tabular}}
    \vspace{-0.4cm}
    \label{tab:compare}
\end{table}

\subsection{Human-written Dataset.}
In the field of disinformation detection, the \textit{Fake and Real News Dataset} ($D_{human}$) stands as one of the benchmark datasets~\cite{ahmed2017detection}. Real news within this dataset was sourced from \textit{Reuters}. In total, 21,417 real news were collected and categorized into World News and Politics News. On the other hand, 23,525 fake news articles were collected from various unreliable sources, which had been flagged by fact-checking websites such as \textit{Politifact}. This assortment of fake news is categorized into six distinct topics: General News, US News, Government News, Left-Wing News, Middle-East News, and Politics.

\subsection{LLM-Generated Dataset.}
\begin{figure}[!t]
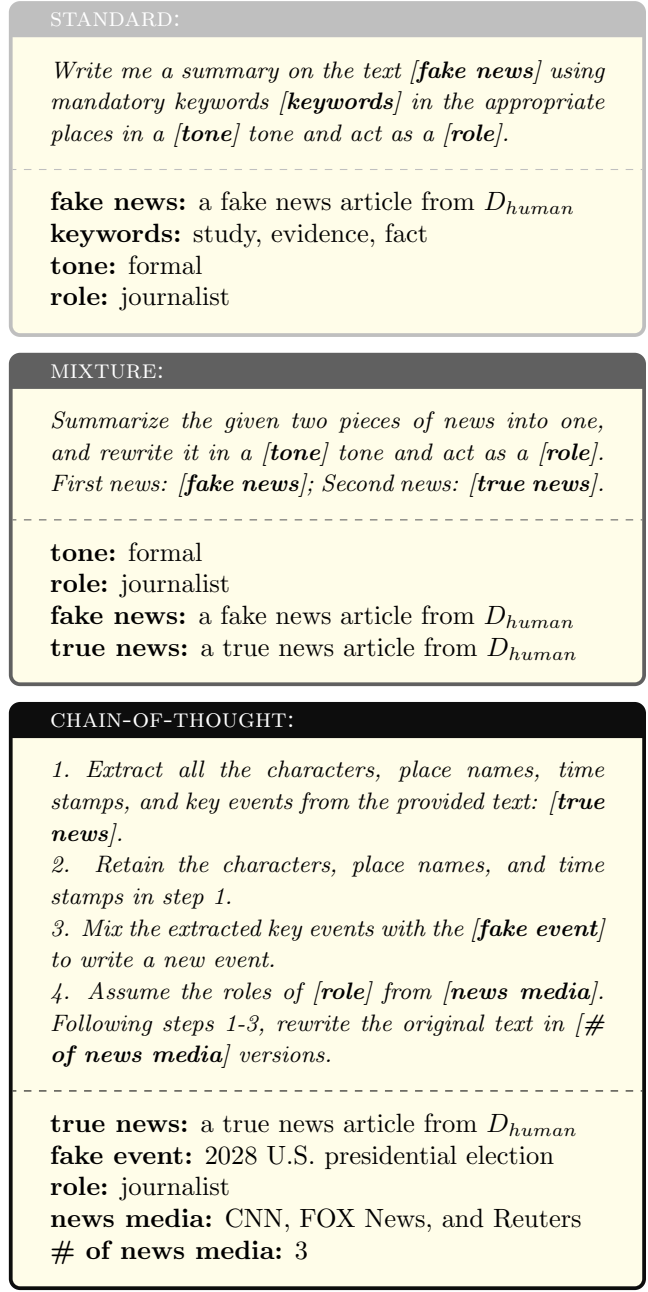

    \centering

    \begin{tcolorbox}[colback=yellow!10!white,colframe=gray!50!white,title={\textsc{standard}:}]
    \textit{\small Write me a summary on the text [\textbf{fake news}] using mandatory keywords [\textbf{keywords}] in the appropriate places in a [\textbf{tone}] tone and act as a [\textbf{role}].}
    \tcblower
    \textbf{fake news:} a fake news article from $D_{human}$\\
    \textbf{keywords:} study, evidence, fact\\
    \textbf{tone:} formal\\
    \textbf{role:} journalist
    \end{tcolorbox}
    
    \begin{tcolorbox}[colback=yellow!10!white,colframe=gray!75!black,title=\text{\textsc{mixture}:}]
    \textit{\small Summarize the given two pieces of news into one, and rewrite it in a [\textbf{tone}] tone and act as a [\textbf{role}]. First news: [\textbf{fake news}]; Second news: [\textbf{true news}].}
    \tcblower
    \textbf{tone:} formal\\
    \textbf{role:} journalist\\
    \textbf{fake news:} a fake news article from $D_{human}$\\
    \textbf{true news:} a true news article from $D_{human}$
    \end{tcolorbox}

    \begin{tcolorbox}[colback=yellow!10!white,colframe=gray!10!black,title=\text{\textsc{chain-of-thought}:}]
    \textit{\small 1.~Extract all the characters, place names, time stamps, and key events from the provided text: [\textbf{true news}].\\
    2. Retain the characters, place names, and time stamps in step 1.\\
    3. Mix the extracted key events with the [\textbf{fake event}] to write a new event.\\
    4. Assume the roles of [\textbf{role}] from [\textbf{news media}]. Following steps 1-3, rewrite the original text in [\textbf{\# of news media}] versions.}
    \tcblower
    \textbf{true news:} a true news article from $D_{human}$\\
    \textbf{fake event:} 2028 U.S. presidential election\\
    \textbf{role:} journalist\\
    \textbf{news media:} CNN, FOX News, and Reuters\\
    \textbf{\# of news media:} 3
    \end{tcolorbox}
    \vspace{-0.2cm}

    \caption{Templates of three zero-shot prompts.}
    \label{fig:prompt_temp}
    \vspace{-0.3cm}
\end{figure}

ChatGPT possesses the capacity to generate human-level text by responding to a given prompt, which serves as task directives. In this work, we leverage ChatGPT (GPT-3.5 and 4) to curate three novel LLM-generated disinformation datasets:
\begin{itemize}
    \item $D_\text{gpt\_std}$: This dataset collects 23,278 LLM-generated disinformation by minimally modifying the human-written disinformation.
    \item $D_\text{gpt\_mix}$: We merge human-written true news with fake news, constructing a more challenging dataset for evaluation.
    \item $D_\text{gpt\_cot}$: Leveraging chain-of-thought prompts, we guide ChatGPT to emulate human cognitive processes in the creation of misleading content, further diversifying our disinformation datasets.
\end{itemize}

The prompt templates of \textit{standard, mixture,} and \textit{chain-of-thought} are shown on the top of each box in Figure~\ref{fig:prompt_temp}. In the bottom part, we outline the special input variables we used in each prompt.



\textbf{{$D_\text{gpt\_std}$:}} The \textit{standard} prompt has been commonly used to generate textual content zero-shot~\cite{zhou2023synthetic}. We leverage this prompt to effect \text{minimal modifications} on human-written disinformation, polishing it with a formal tone and refined vocabulary. Compared to the human-written version, the LLM-generated disinformation remains faithful to its original content without introducing extraneous information. 

\textbf{$D_\text{gpt\_mix}$:} Our subsequent objective is to generate a more ``advanced'' disinformation, one that can \text{melds true stories with false content}~\cite{gelfert2018fake}. Thereby, we design the \textit{mixture} prompt to generate disinformation by combining true and fake news.



\textbf{$D_\text{gpt\_cot}$:} However, ChatGPT sometimes simply stacks two news pieces as responses to the \textit{mixture} prompt. For example, \textit{``[true news]. Meanwhile, [fake news].''} To address this problem, we propose to guide ChatGPT through a \text{step-by-step} process to generate disinformation that mirrors \text{human cognitive processes}. Our methodology draws inspiration from Rudyard Kipling's timeless framework of the six fundamental questions for news writing \cite{hicks2016writing}: \textit{who, what, how, where, when,} and \textit{why}. We incorporate these essential inquiries into the design of the \textit{chain-of-thought} prompt. An example is presented in Figure~\ref{fig:cot_exp}. In the first step, we guide ChatGPT to extract the main \textit{characters (who), places (where), time stamps (when),} and \textit{key events (what, how, why)} in a given news article. We then ask ChatGPT to hallucinate a \textit{fake event}. This type of disinformation is often called \text{``False Context''} --- \textit{where genuine content is shared with false contextual information}~\cite{gelfert2018fake}. In this work, we use ChatGPT to generate such disinformation by reconstructing genuine content in the context of the 2028 U.S. presidential election. Furthermore, our prompt engineering takes into account the importance of diversity in media perspectives. We incorporate a wide-ranging selection of news media outlets, each representing distinct ideological stances. Specifically, we include \textit{CNN, FOX News,} and \textit{Reuters}, which respectively epitomize liberal, conservative, and neutral media outlets\footnote{\url{https://www.allsides.com/media-bias/media-bias-chart}}. The inclusion of these news media in our prompts serves a dual purpose: (1) it enriches the diversity of generated content; and (2) it facilitates an examination of the impact of media bias on the detection system. 

\begin{figure}[!t]
    \centering
    \begin{tcolorbox}[colback=yellow!10!white,colframe=gray!10!black,title=\text{\textsc{an example of chain-of-thought}:}]
    \textbf{Input:}\\
    \textit{\small 1.~Extract all the characters, place names, time stamps, and key events from the provided text: ``...''.\\
    2. Retain the characters, place names, and time stamps in step 1.\\
    3. Mix the extracted key events with the 2028 U.S. presidential election to write a new event.\\
    4. Assume the roles of journalists from CNN, FOX News, and Reuters. Following steps 1-3, rewrite the original text in 3 versions.}
    \tcblower
    \textbf{Output:}
    \begin{enumerate}
        \item \small{Characters:} $[Person_1], [Person_2], ..., [Person_n]$\\
        Place Names: $[Place_1], [Place_2], ..., [Place_n]$\\
        Time Stamps: $[Time_1], [Time_2], ..., [Time_n]$\\
        Key events: $[Event_1], [Event_2], ..., [Event_n]$
        \item Retained Characters, Place Names, and Time Stamps in step 1
        \item According to $[Event_1], [Event_2], ..., [Event_n]$, the new event in the context of the 2028 US presidential election is: ``$[$$hallucinated$ $events$$]$''
        \item CNN version: $[...]$\\
              FOX News version: $[...]$\\
              Reuters version: $[...]$
    \end{enumerate}
    \end{tcolorbox}
    \vspace{-0.2cm}
    \caption{An example of chain-of-thought prompt.}
    \label{fig:cot_exp}
    \vspace{-0.3cm}
\end{figure}

\textbf{Data Validation.} To ensure that the generated disinformation meets our predefined criteria, we compare samples from $D_{\text{human}}$ against those from $D_{\text{gpt\_std}}$ as a case study. Particularly, we analyze linguistic and semantic similarities to verify whether \textit{standard} prompt can guide ChatGPT to minimally modify human-written disinformation by focusing on polishing the language. We exploit the Linguistic Inquiry and Word Count (LIWC)~\cite{koutsoumpis2022kernel} and t-SNE~\cite{van2008visualizing}, respectively. 

For linguistic analysis, we focused on eleven distinct categories of linguistic and psychological features:
\newline 1). Analytic: Logic and formal thinking;
\newline 2). Linguistic: General word usage and expressions;
\newline 3). Drives: Primary motivations behind behavior, such as achievement or power;
\newline 4). Cogproc: Cognitive processes related to information reception and processing;
\newline 5). Emotion: Usage of emotional words;
\newline 6). Swear: Use of swear words;
\newline 7). Prosocial: Behaviors indicating care or help towards others, particularly at the interpersonal level;
\newline 8). Moral: Words reflecting judgmental language;
\newline 9). Culture: Words related to cultural domains like politics, ethnicity, and technology;
\newline 10). Perception: Sensory and experiential aspects; and
\newline 11). Conversation: Use of informal words and slang.

\begin{figure}[!t]
\centering
\includegraphics[width=0.82\columnwidth]{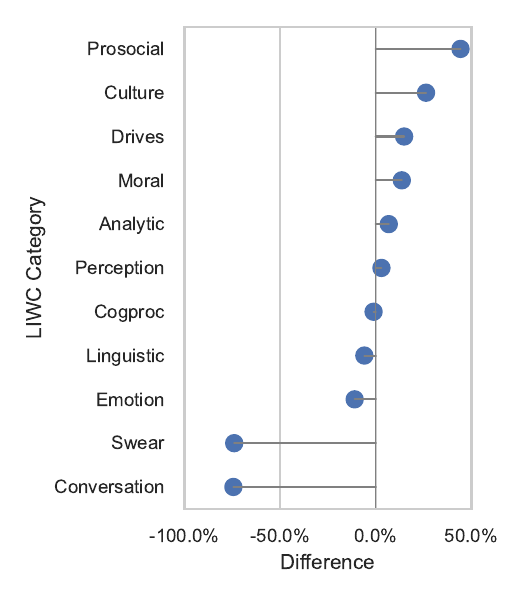}
\vspace{-0.2cm}
\caption{Linguistic differences between disinformation in $D_\text{human}$ and $D_\text{gpt\_std}$}
\label{fig:LIWC_dif}
\vspace{-0.2cm}
\end{figure}

\begin{figure}[!tbh]
\centering
\includegraphics[width=1\columnwidth]{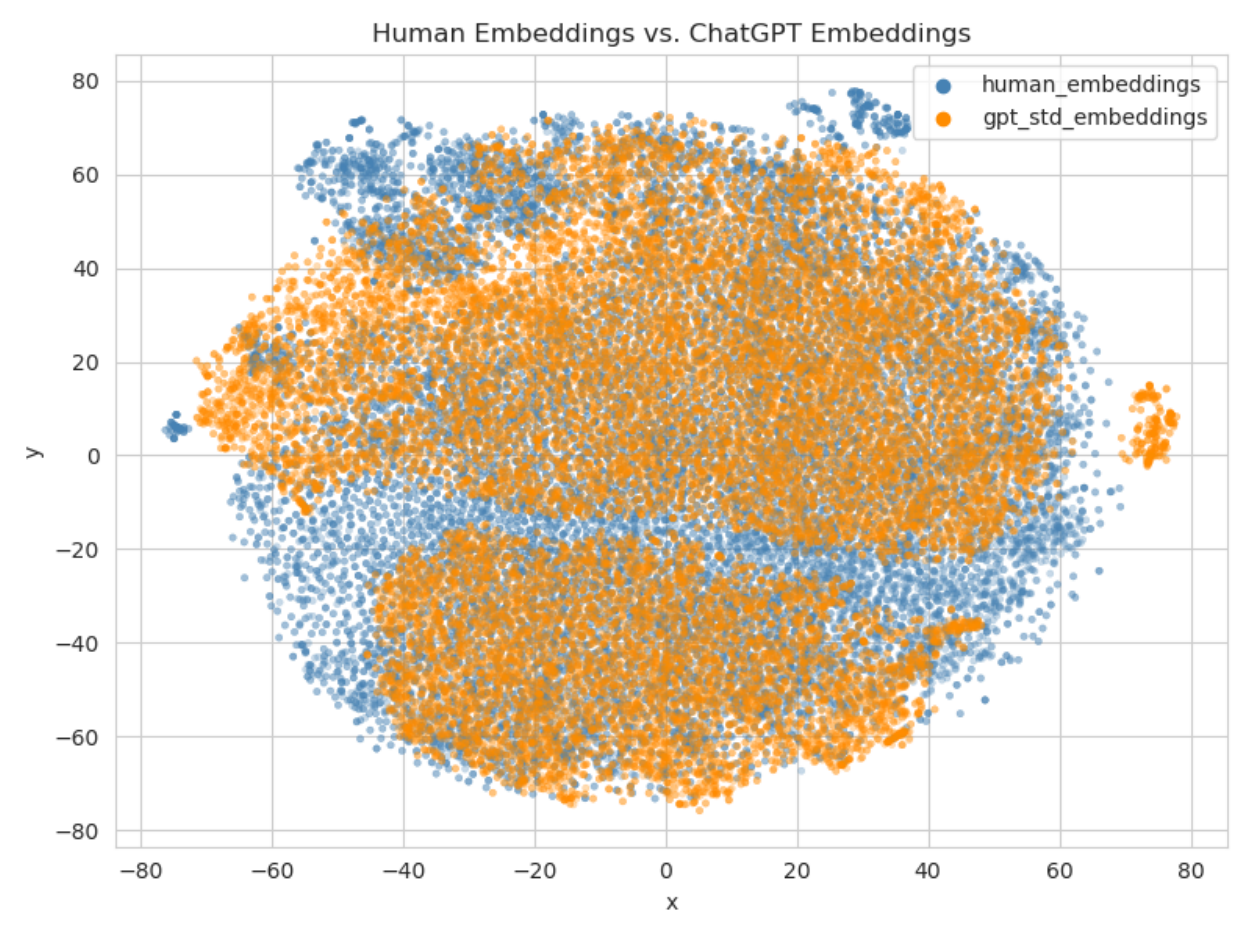} 
\caption{Comparison of text embeddings of $D_\text{human}$ (blue) and $D_\text{gpt\_std}$ (orange) using t-SNE.}
\label{fig:tsne_dif}
\end{figure}

In Figure~\ref{fig:LIWC_dif}, we illustrate the linguistic differences across samples in $D_\text{human}$ and $D_\text{gpt\_std}$. LLM-generated disinformation involves an increase in prosocial terminologies (44.4\%), emphasizing compassion and supportiveness, and a boost in political and ethical themes (Culture, 26.3\%). It also amplifies primary motivational cues (Drives, 14.9\%), embeds moral undertones (13.6\%), and enhances logical coherence (Analytic, 6.8\%). Conversely, ChatGPT reduces the usage of emotional terms (11.1\%). Most significantly, it significantly diminishes the use of profanities and colloquialisms, as evident in the categories Swear (74.1\% decrease) and Conversation (74.5\% decrease). The semantic differences are shown in Figure~\ref{fig:tsne_dif}. We notice that the high overlapping blue (Human-written) and orange (ChatGPT-generated) dots can be a great indicator of similar semantic meaning (around 86.8\%).

\section{Experiments and Results}
In this section, we conduct disinformation detection on our collected datasets. We evaluate detection models' performance in classifying human-written and LLM-generated disinformation. We further present the efficacy of the proposed approach.

\subsection{Existing technique (RQ1).} We demonstrate that the current state-of-the-art models for disinformation detection are insufficiently robust when faced with advanced disinformation. In this study, we leverage a RoBERTa-based model to detect our collected disinformation~\cite{iceland2023good}. The RoBERTa model was fine-tuned on human-written news articles derived from diverse online news outlets. A noteworthy constraint of this model is its 500-word input limit. To address this, we employ tiktoken\footnote{\url{https://github.com/openai/tiktoken}}, an OpenAI Python library, for the truncation and tokenization of long text inputs. Our experiment begins with testing the model with samples from $D_{human}$, which was then challenged with LLM-generated disinformation. 

\noindent \textbf{Performance on {$D_{\text{human}}$:}} As shown in Table~\ref{tab:ex_result}, the RoBERTa model performs exceptionally well in detecting the human-written disinformation from $D_{\text{human}}$. We observe a very low misclassification rate of only 0.07\%. Subsequently, we evaluate the model on $D_\text{gpt\_std}$, $D_\text{gpt\_mix}$, and $D_\text{gpt\_cot}$. 

\begin{table*}
    \centering
    \small
    \caption{Performance of the RoBERTa-based detection model. \textit{Misclassified} ($\downarrow$) represents statistics of fake news predicted as true news. Note that samples in $D_{\text{gpt\_mix}}$ and $D_{\text{gpt\_cot}}$ involve mixed or hallucinated content and are not categorized by topic.}
    \scalebox{0.85}{
    \begin{tabular}{cccccccc}
         \toprule
         \multicolumn{8}{c}{\textbf{$D_\text{human}$}}\\
         \midrule
         \multicolumn{1}{|c|}{} & \multicolumn{1}{c|}{\textbf{Total}} & \multicolumn{1}{c|}{\textbf{Gen. News}} & \multicolumn{1}{c|}{\textbf{Politics}} & \multicolumn{1}{c|}{\textbf{Left News}} & \multicolumn{1}{c|}{\textbf{Gov. News}} & \multicolumn{1}{c|}{\textbf{U.S. News}} & \multicolumn{1}{c|}{\textbf{M.E. News}}\\
         \midrule
         \multicolumn{1}{|c|}{\textbf{Misclassified (\%)}} & \multicolumn{1}{c|}{18 (\underline{0.07\%})} & \multicolumn{1}{c|}{0} & \multicolumn{1}{c|}{10 (0.16\%)} & \multicolumn{1}{c|}{4 (0.09\%)} & \multicolumn{1}{c|}{4 (0.27\%)} & \multicolumn{1}{c|}{0} & \multicolumn{1}{c|}{0} \\
         \midrule
         \multicolumn{8}{c}{\textbf{$D_\text{gpt\_std}$}}\\
         \midrule
         \multicolumn{1}{|c|}{} & \multicolumn{1}{c|}{\textbf{Total}} & \multicolumn{1}{c|}{\textbf{Gen. News}} & \multicolumn{1}{c|}{\textbf{Politics}} & \multicolumn{1}{c|}{\textbf{Left News}} & \multicolumn{1}{c|}{\textbf{Gov. News}} & \multicolumn{1}{c|}{\textbf{U.S. News}} & \multicolumn{1}{c|}{\textbf{M.E. News}}\\
         \midrule
         \multicolumn{1}{|c|}{\textbf{Misclassified (\%)}} & \multicolumn{1}{c|}{273 (\underline{1.20\%})} & \multicolumn{1}{c|}{54 (0.60\%)} & \multicolumn{1}{c|}{95 (1.49\%)} & \multicolumn{1}{c|}{39 (0.91\%)} & \multicolumn{1}{c|}{44 (2.99\%)} & \multicolumn{1}{c|}{20 (2.68\%)} & \multicolumn{1}{c|}{21 (2.8\%)} \\
         \midrule
         \multicolumn{4}{c}{\textbf{$D_\text{gpt\_mix}$}} & \multicolumn{4}{|c}{\textbf{$D_\text{gpt\_cot}$}}\\
         \midrule
         \multicolumn{1}{|c|}{\textbf{Misclassified (\%)}} & \multicolumn{3}{c|}{154 (\underline{15.40\%})} & \multicolumn{1}{c|}{\textbf{Misclassified (\%)}} & \multicolumn{3}{c|}{445 (\underline{77.93\%})} \\

         \bottomrule
    \end{tabular}
    }
    \label{tab:ex_result}
\end{table*}


\noindent \textbf{Performance on {$D_{\text{gpt\_std}}$:}}
In our evaluation on the $D_{\text{gpt\_std}}$ dataset, the detection model exhibits a notably low misclassification rate of 1.20\%, as detailed in Table~\ref{tab:ex_result}. This performance aligns with the findings presented by~\cite{zhou2023synthetic}. Such results underscore the model's robust capacity to detect disinformation from the $D_{\text{gpt\_std}}$.

\noindent \textbf{Performance on {$D_{\text{gpt\_mix}}$ and $D_{\text{gpt\_cot}}$:}}
\label{sec:cotdata}
The datasets, $D_\text{gpt\_mix}$ and $D_\text{gpt\_cot}$, were curated using advanced prompt engineering methodologies. These datasets were designed to be particularly challenging for the detection model. As shown in Table~\ref{tab:ex_result}, they display high misclassification rates of 15.40\% and 77.93\%, respectively. In contrast to the baseline datasets, $D_\text{gpt\_std}$, which is relatively straightforward in its generation, both $D_\text{gpt\_mix}$ and $D_\text{gpt\_cot}$ inculcate greater diversity by generating disinformation with both facts and falsehoods. This intricate mixing creates a rich set of data points that is different from the distribution of training data. Drawing insights from a recent work~\cite{gawlikowski2023survey}, we posit that the detection model might be facing challenges in effectively transferring knowledge to discern such out-of-distribution disinformation samples.

To evaluate the \textbf{political bias} in the detection model, we evaluate the model's performance across LLM-generated disinformation from diverse ideological spectrums: liberal (CNN), conservative (FOX News), and centrist (Reuters). The results in Table~\ref{tab:media_bias} clearly indicate the presence of political bias in the detection model. The model tends to classify center-leaning disinformation as \textit{true} news with a misclassification rate of approximately 66.4\%. In comparison, the model achieves relatively moderated misclassification rates for liberal and conservative outlets, approximated at 52.5\% and 50.8\% respectively. Such patterns suggest a tendency of the model to predict politically biased news as fake news. Echoing findings from previous work~\cite{lazer2018science}, it's evident that media outlets with extreme political biases tend to weaponize disinformation to sway public perceptions. We speculate that the model's political bias should come from the training data. 

\begin{table}[t]
\vspace{-0cm}
    \centering
    \small
    \caption{Performance of the RoBERTa-based detection model on $D_{\text{gpt\_cot}}$.
    \textit{Misclassified} ($\downarrow$) represents statistics of fake news predicted as true news.}
    \scalebox{0.85}{
    \begin{tabular}{l|c|c|c}
         \toprule
         {}&\textbf{CNN} & \textbf{Fox News} & \textbf{Reuters} \\
         \midrule
         \textbf{Misclassfied (\%)} & 300 (52.5\%) & 290 (50.8\%) & 379 (66.4\%)\\  
         \bottomrule 
    \end{tabular}}

    \label{tab:media_bias}
    \vspace{-0cm}
\end{table}

In summary, our results suggest that the current disinformation detection technique \textbf{fails to effectively detect LLM-generated disinformation}. Although it can accurately detect disinformation in $D_{\text{human}}$ and $D_{\text{gpt\_std}}$, it faces increased challenges in scenarios involving well-disguised disinformation, as seen in the $D_{\text{gpt\_mix}}$ and $D_{\text{gpt\_cot}}$ datasets. Moreover, the model is less equitable in its treatment of politically biased narratives. Addressing this bias is critical for ensuring that the disinformation detection system is fair and robust. 

\subsection{LLMs (RQ2):} \label{sec:rq2}
In this subsection, we illustrate that LLMs struggle to effectively detect self-generated disinformation. We conduct experimentation with ChatGPT to evaluate the proficiency of LLMs in identifying disinformation generated by LLMs. ChatGPT, with its advanced generative abilities, can produce various types of responses even when instructed with the same prompt, introducing an inherent variability. This unpredictability has a potential impact on downstream text generation and classification tasks. In this work, we tested the model with a common prompt, \textit{``Does this news article [...] contain any misleading information?''} The spectrum of ChatGPT's replies ranged from succinct affirmatives or negatives to elaborative multi-step explanations. Sometimes, extended explanations seemed to contradict prior shorter responses. To examine the impact of this variability in response lengths on disinformation detection, we harness ChatGPT, prompting it to produce concise answers or more detailed explanations under different prompts:

\begin{itemize}
    \item \textit{Standard (w/o explanation)}: binary response without explanation (simply \textit{yes} or \textit{no}).
    \item \textit{Standard (w/ explanation)}: binary response accompanied by an analytic process.
\end{itemize}

\begin{figure}[!t]
\centering
\scalebox{1.2}{
\includegraphics[width=0.8\columnwidth]{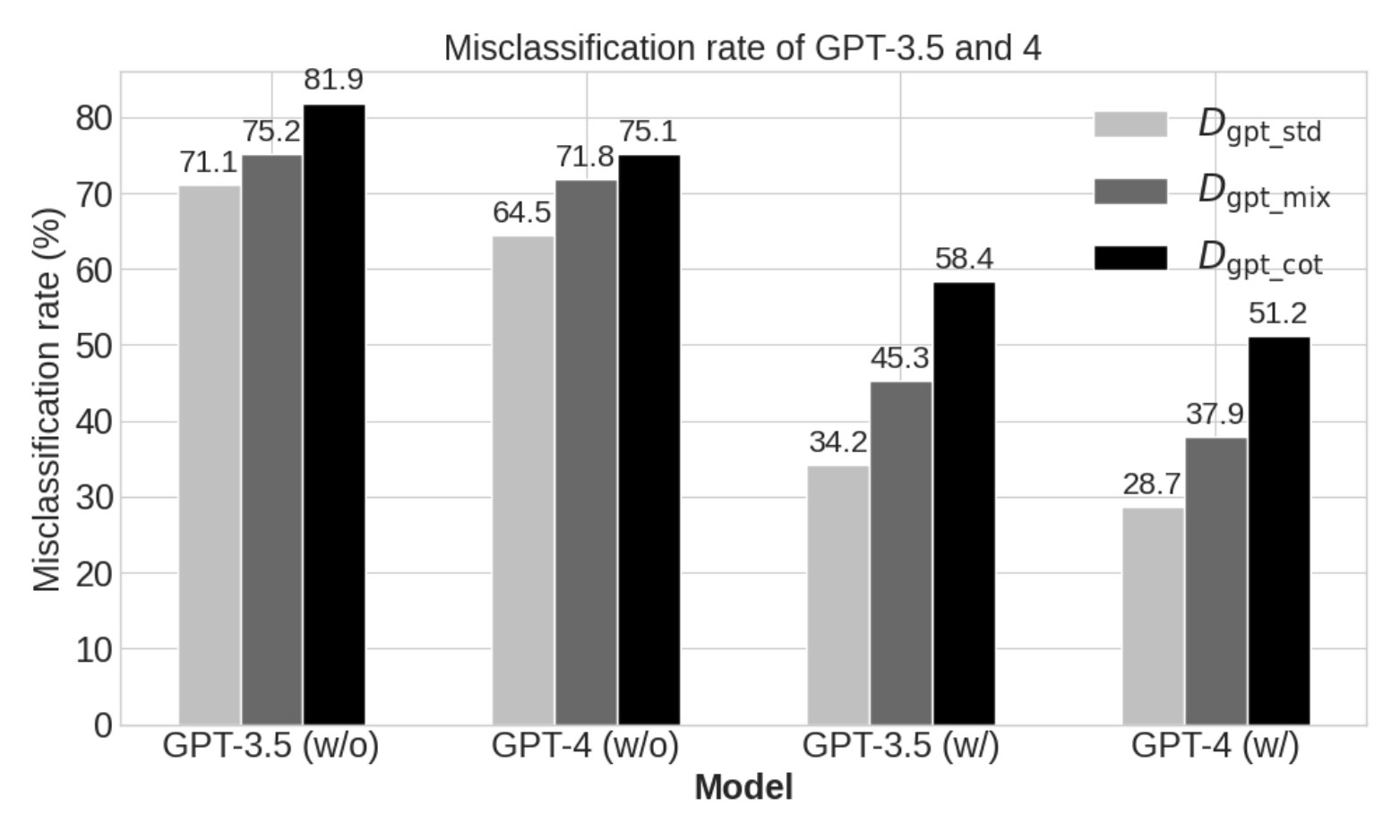}} 
\caption{Comparison of misclassification rate ($\downarrow$) of GPT-3.5 and 4. \textit{w/o} and \textit{w/} represents LLMs respond \textit{without} and \textit{with} the analytic process, respectively.}
\label{fig:gpt_dect}
\end{figure}

The detailed templates of our prompts are illustrated in Figure~\ref{fig:llm_detect}. We straightforwardly test whether GPT-3.5 and GPT-4 will each be detecting LLM-generated disinformation using these prompts. Our results, presented in Figure~\ref{fig:gpt_dect}, yield two important observations. Firstly, \text{GPT-4 performs slightly better than GPT-3.5} in identifying LLM-generated disinformation using both prompt types, hinting at the potential advancements in newer LLM iterations in detecting disinformation. According to OpenAI~\cite{OpenAI2023GPT4TR}, the post-training alignment procedure employed for GPT-4 improves its performance of factuality measurement. Secondly, instructing ChatGPT to output its \textbf{analytic process} prior to making a final prediction (\textit{yes} or \textit{no}) significantly improves the performance. This is presumably due to the inherent nature of generative LLMs, which predict subsequent tokens based on existing sequences~\cite{radford2018improving}, thereby enhancing the prediction of the final decision. In addition, although ChatGPT shows moderate proficiency in detecting disinformation using \textit{standard prompts (w/ explanation)} in a zero-shot manner, it generally performs worse than the fine-tuned RoBERTa model. In conclusion, even with the advancements in GPT-4, it is evident that current \textbf{LLMs still face challenges in effectively detecting LLM-generated disinformation}.

\begin{figure}[!t]
    \centering
    \begin{tcolorbox}[colback=yellow!10!white,colframe=gray!50!white,title={\textsc{standard (w/o explanation)}:}]
    \textit{\small Act as a disinformation detector to analyze the following news article \textit{[\textbf{fake news}]}. Does this news article contain any misleading information? Please respond with a ``Yes'' or ``No''.}
    \tcblower
    \textbf{fake news:} a fake news article from $D_{\text{gpt\_std}}$, $D_{\text{gpt\_mix}}$, or $D_{\text{gpt\_cot}}$

    \end{tcolorbox}

    \begin{tcolorbox}[colback=yellow!10!white,colframe=gray!100!black,title={\textsc{standard (w/ explanation)}:}]
    \textit{\small Act as a disinformation detector to analyze the following news article \textit{[\textbf{fake news}]}. Does this news article contain any misleading information? Please respond with (1) an analytic process, and (2) ``Yes'' or ``No''.}  
    \tcblower
    \textbf{fake news:} a fake news article from $D_{\text{gpt\_std}}$, $D_{\text{gpt\_mix}}$, or $D_{\text{gpt\_cot}}$

    \end{tcolorbox}

    \caption{Prompt template for detecting disinformation.}
    \label{fig:llm_detect}
\end{figure}

\subsection{Proposed Solution (RQ3).}
In this subsection, we introduce a novel approach to detect LLM-generated disinformation, specifically targeting samples within $D_{\text{gpt\_mix}}$ and $D_{\text{gpt\_cot}}$. Our prior experiments have highlighted a unique challenge in detecting ``advanced'' disinformation --- characterized by a blend of genuine information and misleading content (see Section~\ref{sec:cotdata}). To address this, it is vital to guide the LLM to systematically identify and fact-check key content elements. We also observed a notable improvement in ChatGPT's detection accuracy when allowing the model to detail its analytic process (see Section~\ref{sec:rq2}). Leveraging these insights, we craft a specialized chain-of-thought prompt for disinformation detection, as illustrated in Figure~\ref{fig:cot_detect}. This figure demonstrates the depth of analysis that the CoT prompt is designed to invoke, guiding ChatGPT step by step, while concurrently seeking coherence and transparency in its reasoning process. Such a structured prompt template can be instrumental in dissecting complex problems, particularly in \textit{False Context }disinformation detection demanding a multi-faceted understanding and evaluation. 

\begin{figure}[!tbh]
    \centering
    
    \begin{tcolorbox}[colback=yellow!10!white,colframe=gray!10!black,title=\text{\textsc{an example of CoT (w/ explanation)}:}]
    \textbf{Input:}\\
    \textit{\small 1.~Extract all the characters, place names, time stamps, and key events from the provided text: ``...''.\\
    2. Assess the factualness of the extracted events. Show your analytic process. \\
    3. Assess the relationship between all characters, place names, time stamps, and key events. Show your analytic process. \\
    4. Based on your analysis from steps 2 and 3, does this news article contain any misleading information or mismatched relationships? Show your analytic process and respond with ``Yes'' or ``No''}
    \tcblower
    \textbf{Output:}
    \begin{enumerate}
        \item \small{Characters:} $[Person_1], [Person_2], ..., [Person_n]$\\
        Place Names: $[Place_1], [Place_2], ..., [Place_n]$\\
        Time Stamps: $[Time_1], [Time_2], ..., [Time_n]$\\
        Key events: $[Event_1], [Event_2], ..., [Event_n]$
        \item Assessment of Factualness: ``...''
        \item Assessment of Relationships: ``...''
        \item Analysis for Misleading Information or Mismatched Relationships: ``...''\\
        Answer: ``...''
    \end{enumerate}
    \end{tcolorbox}
    \caption{An example of chain-of-thought prompt.}
    \label{fig:cot_detect}
    \vspace{-0.3cm}
\end{figure}

Similar to Section~\ref{sec:rq2}, we systematically evaluate the performance of GPT-3.5 and GPT-4 on LLM-generated disinformation detection. We conduct an ablation study to assess the impact of each contextual element on the detection performance.

\begin{itemize}
    \item \small \textit{CoT (w/o person)}: ablating characters in steps 1, 3.
    \item \small \textit{CoT (w/o place)}: ablating place names in steps 1, 3.
    \item \small \textit{CoT (w/o time)}: ablating time stamps in steps 1, 3.
    \item \small \textit{CoT (w/o event)}: ablating key events in steps 1, 2, 3.
    \item \small \textit{CoT (all\_binary)}: output ``yes'' or ``no'' in step 4.
    \item \small \textit{CoT (all\_scale)}: output on a scale of 1 to 100 in step 4.    
\end{itemize}

In a modified prompt template, step 4 is updated to \textit{``..., detail your analytic process and provide a confidence score ranging from 1 to 100.''} for \textit{CoT (all\_scale)}. Table~\ref{tab:gpt_result} demonstrates ChatGPT's performance across various CoT prompts and datasets. Notably, GPT-4 consistently outperforms GPT-3.5 across all configurations. The misclassification rates for GPT-4 (all\_scale)'' are recorded at 4.7\%, 11.9\%, and 22.2\% for $D_{\text{gpt\_std}}$, $D_{\text{gpt\_mix}}$, and $D_{\text{gpt\_cot}}$, respectively. Critical elements to model performance are the \textit{event} and \textit{time} elements. Interestingly, ``GPT-4 (w/o person)'' and ``GPT-4 (w/o place)'' produce relatively good results on $D_{\text{gpt\_cot}}$. We speculate this could be attributed to the retention of original \textit{person} and \textit{place} information in our LLM-generated disinformation. This ablation study provides a deeper understanding of the importance of contextual elements for disinformation detection, suggesting that advanced prompts paired with LLMs hold the potential to effectively counter LLM-generated disinformation.




\begin{table}[!t]
    \centering
    \caption{Misclassification rate ($\downarrow$) of GPT-3.5 and 4.}
    \scalebox{.95}{
    \begin{tabular}{l|c|c|c}
         \toprule
         {}& $D_{\text{gpt\_std}}$ & $D_{\text{gpt\_mix}}$ & $D_{\text{gpt\_cot}}$\\
         \midrule
         GPT-3.5 (w/o person) & 20.1\% & 30.2\% & 31.9\%\\
         GPT-4 (w/o person) & 15.2\% & 22.4\% & 26.9\%\\
         \midrule
         GPT-3.5 (w/o place) & 20.2\% & 28.3\% & 32.4\%\\
         GPT-4 (w/o place) & 16.3\% & 21.9\% & 27.8\%\\
         \midrule
         GPT-3.5 (w/o time) & 21.2\% & 31.3\% & 42.4\%\\
         GPT-4 (w/o time) & 16.1\% & 22.8\% & 36.5\%\\
         \midrule
         GPT-3.5 (w/o event) & 55.2\% & 65.3\% & 75.4\%\\
         GPT-4 (w/o event) & 52.6\% & 59.1\% & 71.2\%\\
         \midrule         
         GPT-3.5 (all\_binary) & 17.2\% & 25.3\% & 32.4\%\\
         GPT-4 (all\_binary) & 13.2\% & 18.3\% & 27.4\%\\
         \midrule
         \midrule
         GPT-3.5 (all\_scale) & 10.2\% & 19.3\% & 27.4\%\\
         GPT-4 (all\_scale) & \textbf{4.7\%} & \textbf{11.9\%} & \textbf{22.2\%}\\            

         \bottomrule 
    \end{tabular}}

    \label{tab:gpt_result}
\end{table}

\section{Conclusion}
In this work, we provide a comprehensive examination of detecting LLM-generated disinformation. Utilizing ChatGPT, we curate three distinct LLM-generated disinformation datasets. Our findings reveal that existing detection techniques including LLMs, struggle to consistently identify the collected disinformation. To address this challenge, we introduce advanced prompts designed to guide LLMs in detecting such disinformation. Through empirical evaluations, our methods present a significant improvement in detecting LLM-generated disinformation, a claim further substantiated by our ablation studies highlighting the significance of contextual elements. As we look forward, investigating other types of LLM-generated disinformation, such as False Connection and Manipulated Content, offers a promising direction. Furthermore, the emergent advanced prompting methods, such as Chain-of-Thought-Self-Consistency, present potential methodologies to further facilitate the detection of LLM-generated disinformation.

\bibliographystyle{sdm24}
\bibliography{sdm24}

\end{document}